\documentclass[letterpaper, 10 pt, conference]{ieeeconf}

\IEEEoverridecommandlockouts

\usepackage{graphicx}

\title{\LARGE \bf
A Multi-Operator Mixed-Reality Interface for Multi-Robot Control and Coordination: Co-Located and Private Workspace Collaboration
}

\author{Omotoye Shamsudeen Adekoya*, Antonio Sgorbissa, Carmine Tommaso Recchiuto% <-this % stops a space
\thanks{* Corresponding author: omotoye.adekoya@edu.unige.it{\tt\small }} % <-this % stops a space
\thanks{All authors are with DIBRIS Department, RICE Laboratory, University of Genoa, Italy
}
}

\begin{document}

\maketitle
\thispagestyle{empty}
\pagestyle{empty}

%%%%%%%%%%%%%%%%%%%%%%%%%%%%%%%%%%%%%%%%%%%%%%%%%%%%%%%%%%%%%%%%%%%%%%%%%%%%%%%%
\begin{abstract}
Multi-operator control of robot teams requires not only access to the same mission information, but also mechanisms for maintaining shared awareness and preventing conflicting interventions. Building on our previous HORUS interface \emph{(Holistic Operational Reality for Unified Systems)} we present a mixed-reality interface that extends single-operator multi-robot supervision to collaborative multi-operator use. The system supports two complementary modes: a co-located shared workspace, in which operators observe and manipulate the same mini-map in the same physical location, and a private-workspace mode, in which operators work on the same mission through independently placed local workspaces. The architecture combines registration-driven scene construction, lightweight shared-session synchronization, and per-robot control leases to support collaborative monitoring, tasking, and teleoperation while preventing conflicting commands. We evaluated the approach in a human-subject study with 36 participants (18 pairs) controlling three Nova Carter mobile robots in two search environments. The performance of the objective task was comparable across the two modes, indicating that both modes supported effective mission execution. However, the co-located shared workspace significantly improved perceived collaboration, shared understanding, and handoff clarity, and was the preferred collaborative mode. These results indicate that physically co-locating the MR workspace improves how operators coordinate even when the underlying robot-control tools remain unchanged.
\end{abstract}

\section{Introduction}
\label{sec:introduction}

Multi-robot systems are increasingly deployed in domains such as search and rescue, industrial inspection, and exploration, demanding time-critical decisions and accurate situational awareness across multiple agents. However, scaling from single-robot to team-level operation remains challenging: conventional 2D ground control stations (GCS) force operators to mentally integrate maps, camera feeds, and robot states. This cognitive fusion degrades awareness and increases workload as mission complexity increases \cite{Chen2011SupervisoryControlReview,Lewis2013MultipleRemoteRobots,Kolling2016HSISurvey}. Consequently, the human operator often becomes the bottleneck for understanding team state and intervening effectively.

Immersive and mixed-reality (MR) interfaces aim to reduce this burden by presenting the state of the robot and the spatial context in three-dimensional embodied ways \cite{Roldan2017MultiRobotInterfacesSA,Frank2017TowardMRMultiRobot,Suzuki2022AugmentedRealityRobotics}. Recent systems show that spatial interaction metaphors can improve multi-robot task allocation and planning compared to 2D tools \cite{KennelMaushart2023MultiRobotMR,Chen20243DMixedReality}. Our prior work, HORUS (Holistic Operational Reality for Unified Systems), introduced an MR mini-map workspace integrating multi-robot monitoring, task-assignment, and single robot teleoperation within a unified exocentric view \cite{Adekoya2025HORUS}. While these systems substantially improve \emph{single-operator} management, realistic missions often require multiple humans to collaborate by dividing attention, negotiating priorities, and handing off control.

Supporting such collaboration is both an interface and a coordination challenge. Effective collaboration requires mutual awareness of teammate's actions, coupled with mechanisms preventing conflicting commands and unsafe interference \cite{Rule2012MultiUserMultiRobotInterfaces,MusicHirche2017ControlSharing,Miyauchi2023SharingSwarmControl}. Enabling this in \emph{co-located} mixed reality introduces additional constraints: multiple headsets must share a common spatial frame so all operators perceive the identical virtual workspace in the same physical location, robust to drift and cross-user inconsistency \cite{McGill2020QuestCoLocatedMR,Ran2020SpatialConsistencyCoNEXT,Dhakal2022SLAMShare}. Moreover, collaborative control demands explicit authority management so task assignments and teleoperation inputs coordinate rather than compete \cite{Noohi2015DynamicAuthorityIROS,Liu2015DualUserAuthorityEMBC,Lu2017DualUserTransparency,Fern2018MOMU}.

\begin{figure}[t]
    \centering
    \includegraphics[width=\columnwidth]{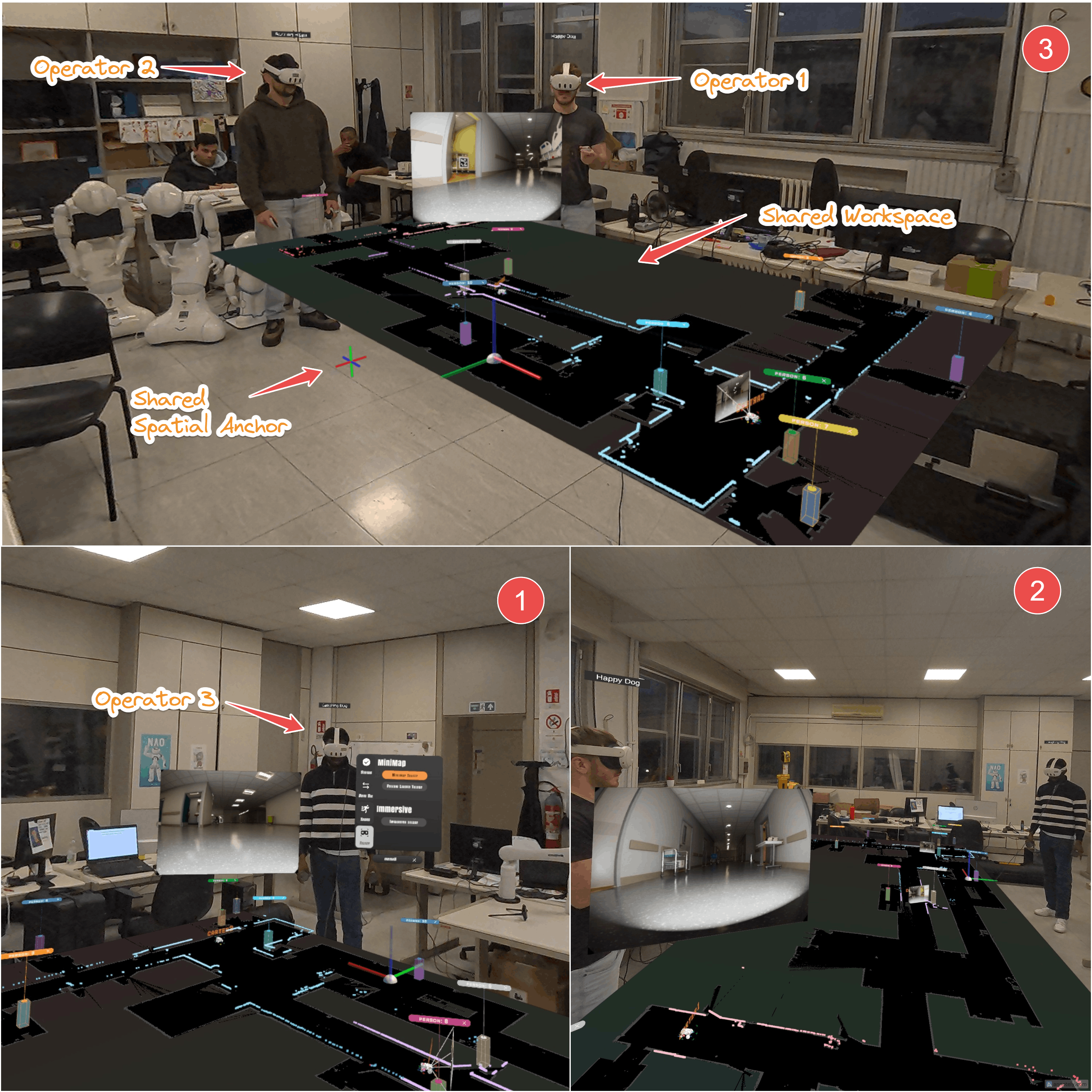}
    \caption{HORUS in co-located shared-workspace mode. 3 operators observe and interact with the same spatially anchored miniature workspace from different viewpoints. The number label indicates the operator number whom the point of view belongs to.}
    \label{fig:shared_workspace_three_operators}
\end{figure}

In this paper, we present a mixed-reality interface extending HORUS to \emph{multi-operator} multi-robot control and coordination. The system supports collaborative sessions across headsets via two complementary modes: a \emph{co-located shared-workspace} mode, where operators observe the same mini-map in the same physical location, and a \emph{private-workspace} mode, where operators participate in the same mission via independently placed local workspaces. This paper explicitly isolates the effect of shared physical co-location through a controlled human-subject study.

Our main contributions are:
\begin{itemize}
    \item a multi-operator MR architecture supporting both co-located and private-workspace collaboration via shared-state synchronization;
    \item a workspace-centric interaction design featuring explicit mechanisms for shared tasking, awareness cues, and conflict prevention;
    \item a dynamic control arbitration and handoff approach utilizing per-robot leases;
    \item an evaluation combining system-level validation with a human-subject study comparing co-located shared-workspace against private-workspace collaboration.
\end{itemize}

\section{Related Work}
\label{sec:related_work}

\subsection{Mixed Reality Interfaces for Multi-Robot Control}
MR interfaces improve multi-robot monitoring by reducing indirection and presenting spatial relationships naturally \cite{Frank2017TowardMRMultiRobot,Roldan2017MultiRobotInterfacesSA}. Systems supporting 3D tasking report benefits for planning compared to 2D interfaces \cite{KennelMaushart2023MultiRobotMR,Chen20243DMixedReality}. Other works emphasize interaction patterns for heterogeneous teams and in-situ visualizations of navigation paths \cite{ChaconQuesada2020ARHeteroMultirobot}, while HORUS integrated team monitoring, task-assignment tools, and teleoperation into an MR mini-map paradigm \cite{Adekoya2025HORUS}. The present work extends this from single-operator management to multi-operator collaboration.

\subsection{Multi-Operator Collaboration and Authority Management}
Multi-user interface research argues that collaboration requires explicit support for mutual awareness, division of labor, and coordination \cite{Rule2012MultiUserMultiRobotInterfaces}. While sharing swarm control improves parallelism, it introduces coordination costs and conflicting command risks \cite{Kolling2016HSISurvey,Miyauchi2023SharingSwarmControl}. A recurring issue is \emph{authority management}. Prior work in dual-user teleoperation highlights strategies for dynamic authority distribution and dominance transparency \cite{Noohi2015DynamicAuthorityIROS,Liu2015DualUserAuthorityEMBC,Lu2017DualUserTransparency,MusicHirche2017ControlSharing}. Similar structured handoff strategies improve safety in multi-vehicle settings \cite{Fern2018MOMU}. These findings motivate our explicit arbitration mechanisms for collaborative MR multi-robot control, rather than assuming collaboration emerges from shared visualization alone.

\subsection{Co-Located Multi-User MR Alignment and Consistency}
Co-located MR requires headsets to agree on a common spatial frame. Research highlights failure modes like cross-user spatial inconsistency and latency \cite{McGill2020QuestCoLocatedMR,Ran2020SpatialConsistencyCoNEXT,Dhakal2022SLAMShare}. Practical shared-anchor building blocks are increasingly enabling markerless same-space MR workflows \cite{He2021SpatialAnchorIndoorAssetTracking,VidalBalea2025OffCloudAnchors}. Despite these advances, co-located alignment is rarely evaluated in robot fleet command-and-control contexts. Our work targets this gap by comparing shared physical MR alignment against a private-workspace alternative.

\section{System and Interaction}
\label{sec:system}

\begin{figure*}[!t]
\centering
\IfFileExists{figures/system_architecture.png}{%
\includegraphics[width=1\textwidth]{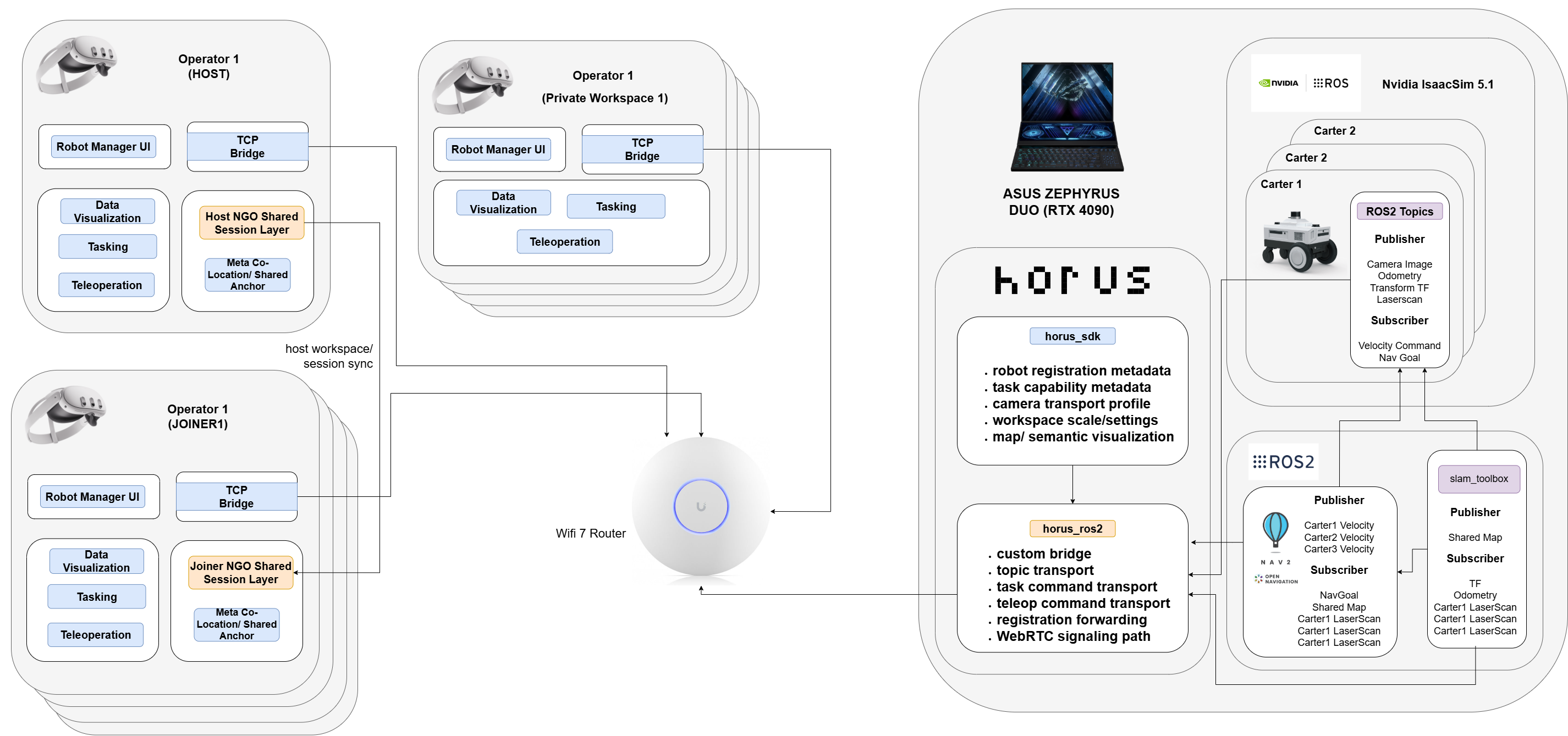}%
}{%
}
\caption{System Architecture Diagram}
\label{fig:system_arch}
\end{figure*}

The proposed system extends HORUS from a single-operator MR interface to a collaborative multi-operator runtime while preserving the same workspace-centric interaction model \cite{Adekoya2025HORUS}. Instead of relying on disconnected 2D panels, operators reason about the robot team through a spatially registered miniature workspace. In the multi-operator setting, this workspace becomes the common reference for monitoring, tasking, and intervention, while the system prevents conflicting commands on the same robot. Although the controlled study focuses on pairs, the runtime is not restricted to two users.

At runtime, HORUS maintains a miniature exocentric model of the remote robot environment. Robots, maps, sensor overlays, paths, labels, camera panels, and control widgets are anchored to a common workspace. Operators can inspect the scene from different viewpoints, open per-robot control panels, assign tasks directly on the map, and switch from supervision to teleoperation without leaving the MR scene. The multi-operator extension adds two capabilities: support for multiple collaborative workspace modes and per-robot arbitration for conflict-safe parallel work.

\subsection{Runtime Architecture}
HORUS follows a three-layer architecture (see Fig. \ref{fig:system_arch}). The SDK layer publishes robot registration metadata, maps, sensors, and task capabilities. The ROS~2 bridge handles command, state, and map communication, while immersive video can use WebRTC. The Unity MR runtime instantiates the workspace, visualizations, teleoperation interfaces, and multi-operator coordination logic. A key design decision is to separate \emph{robot-state transport} from \emph{multi-user MR synchronization}: robot data remain in the robotics transport layer, while low-rate collaborative state is synchronized separately through Unity Netcode for GameObjects (NGO).

\subsection{Registration-Driven Workspace Construction}
The runtime does not assume a fixed set of robots or streams. Instead, it dynamically constructs the scene from SDK registration payloads. For each active robot, HORUS instantiates a spatial anchor, an interaction surface, a Robot Manager panel, task controls, and camera logic.

Operators first create or enter a workspace and then accept it, defining the session's operational coordinate frame. The workspace is layered into a floor grid, map layer, and geometric overlays for paths and task markers. Workspace scale is configurable so that robot poses, goals, and interaction distances remain consistent in both co-located and private-workspace operation.

\begin{figure}[t]
    \centering
    \includegraphics[width=\columnwidth]{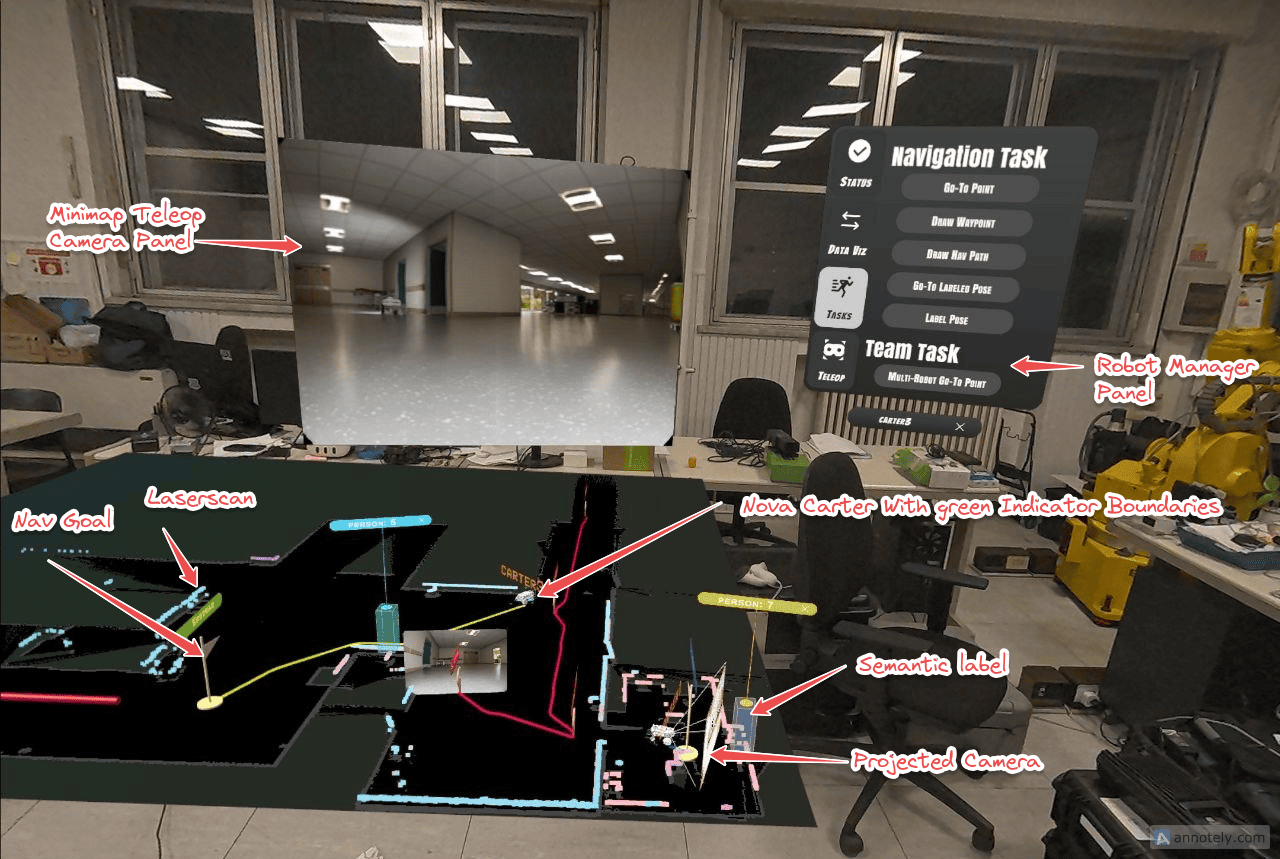}
    \caption{Workspace-centric interaction in HORUS. The operational workspace contains robots, maps, overlays, and task markers anchored in a common frame.}
    \label{fig:workspace_interaction}
\end{figure}

\subsection{Monitoring, Tasking, and Teleoperation}

Each robot is represented by a spatial anchor, a body visualization, and a Robot Manager panel. Robot pose is computed by running a collaborative mapping software (slam\_toolbox MultiSLAM) and extracted through the ROS~2 bridge interface, and HORUS can render simplified or description-driven bodies, collision models, joint axes, goals, waypoints, and planned paths. Selecting a robot opens its Robot Manager and makes it the focus of subsequent actions.

The Robot Manager is the per-robot entry point for both supervision and intervention. From it, operators can inspect robot state, enable or disable visualization channels, start teleoperation, and author tasks. HORUS combines workspace-level visualizations with robot-specific information. Workspace-level layers include occupancy maps, point clouds, meshes, and semantic overlays, while per-robot channels include camera feeds, scans, plans, waypoint queues, trails, and collision-related overlays. This allows operators to filter information in dense multi-robot scenes.

Task authoring is performed directly inside the workspace. For ground robots, HORUS supports actions such as navigating to a point in the environment, following a given path, labelling a pose, or multi-robot navigation to a point in the environment. Goals and paths are placed and manipulated directly in the miniature environment, keeping task specification spatially grounded.

HORUS supports multiple teleoperation modalities. In minimap teleoperation, the operator maintains the exocentric overhead view while sending motion commands. In immersive teleoperation, the operator receives a more direct camera-centric view and non-selected camera streams can be suppressed to reduce clutter. Supervisory views typically remain on ROS transport, whereas immersive teleoperation can switch to WebRTC when lower latency is required. In collaborative sessions, teleoperation panels are placed so they do not obscure the main workspace.

\subsection{Collaborative Modes and Shared Spatial Alignment}

The runtime supports two collaborative multi-operator modes. In \emph{co-located shared-workspace mode}, all participants observe the same workspace in the same physical location. In \emph{private-workspace mode}, participants collaborate on the same mission and share operational state, but each works from an independently placed local workspace. The user study compares these two modes to isolate the effect of physical workspace co-location.

Both modes support more than two participants. The shared-session and lease layers are not pair-specific.

In private-workspace mode, each operator creates a local workspace, while robot registrations, maps, task state, and lease state remain shared through the common backend. In shared host/join mode, one participant creates the workspace and advertises the session, while other participants discovers and join the session, localize the shared spatial anchor, and then explicitly join the workspace. HORUS relies on Meta's co-location building blocks and shared spatial anchors to establish the common frame. To avoid displaying an unaligned state, the system defers loading the robots and maps until the user successfully localizes the shared anchor and explicitly clicks to join the workspace.

Collaborative synchronization is intentionally narrow in scope. NGO custom messages exchange low-rate application state, including workspace snapshots, registration boundaries, deltas, unregister events, and resync requests. Rather than continuously synchronizing every Unity object over the network, the SDK resends the initial robot and map registration data to the joining operator, allowing their local headset to reconstruct the scene. This keeps the robotics-side registration model as the source of truth and makes late-join recovery more robust.

% \begin{figure}[t]
%     \centering
%     % \includegraphics[width=\columnwidth]{figures/shared_session_flow.pdf}
%     \fbox{\parbox[c][4.0cm][c]{0.95\columnwidth}{\centering Placeholder for collaborative modes and lease workflow figure}}
%     \caption{Collaborative operation in HORUS. The system supports both co-located shared-workspace sessions and collaborative private-workspace sessions, while lease-aware gating prevents conflicting actions on the same robot and preserves parallel work on different robots.}
%     \label{fig:shared_session_flow}
% \end{figure}

\subsection{Conflict-Safe Parallelism Through Per-Robot Leases}

Shared mission participation alone is not sufficient for collaborative robot control: the system must also prevent conflicting commands. HORUS addresses this through \emph{per-robot control leases}. Leases are enforced at the robot level rather than at the whole-workspace level, allowing multiple operators to work in parallel on different robots while preventing simultaneous active control of the same one.

A lease is associated with an action context such as opening a robot panel, starting task authoring, or starting teleoperation. Lease state tracks whether the holder is actively using the robot through three activity flags: \texttt{panel\_open}, \texttt{task\_active}, and \texttt{teleop\_active}. A robot is unavailable only when another operator holds the lease and at least one of these flags indicates active use. If a holder is no longer active, HORUS treats the lease as stale and permits reacquisition.

The lease protocol uses dedicated multi-operator topics for acquire, heartbeat, and release messages. The policy is simple: all participants have equal priority; the first active claimant holds the robot; and inactive holders can be displaced through reacquisition. Before starting teleoperation or tasking, the runtime checks whether the current operator is allowed to act. If another operator actively holds the lease, the action is blocked; if the lease is stale, HORUS reacquires the robot before proceeding.

To make this authority model legible, each robot exposes an availability state directly in the workspace and Robot Manager: \emph{white} indicates that the robot is free, \emph{green} indicates that it is currently held by the local operator, and \emph{red} indicates that it is held by another operator. The same state is reflected functionally in the interface. If a robot is red, robot-specific actions that would create a conflict, such as teleoperation or task assignment, are shown as unavailable to the other operator until the lease is released or becomes inactive. When the robot returns to white, those actions become available again. This makes the arbitration model legible both visually and functionally: operators can identify robot availability at a glance, and the interface prevents conflicting commands at the point of interaction.

In both collaborative modes, this supports conflict-safe parallelism: operators can work independently on different robots, but actions on the same robot remain coordinated through explicit arbitration. This mechanism may help operators collaborate even when they are not in the same location and cannot communicate directly with one another.

\section{Experimental Design}
\label{sec:experimental_design}

We evaluated whether physically co-locating the MR workspace changes how pairs of operators coordinate a mission compared to private-workspace collaboration. We hypothesized that co-location would improve subjective coordination, intent awareness, and collaborative effectiveness by providing a mutually observable spatial reference. We also hypothesized that the private-workspace mode would be reliable enough to allow multiple operators to perform the task even without direct communication. Although HORUS has already been validated in prior real-robot experiments \cite{Adekoya2025HORUS}, the present study was simulation-based to provide repeatable, safe, and controlled conditions while preserving the full MR interaction loop.

Based on these hypotheses, we formulate the following research questions:
\begin{itemize}
    \item \textbf{RQ1:} Does physical co-location of a shared MR workspace improve perceived collaboration and shared awareness compared to a private-workspace setup?
    \item \textbf{RQ2:} Can operators maintain effective multi-robot task performance in a private MR workspace without direct verbal communication?
\end{itemize}

\subsection{Apparatus and Simulated Setup}

The study used two Meta Quest~3 headsets, one per participant, running the HORUS runtime described in Section~\ref{sec:system}. Participants interacted through standard Quest controllers. The simulation and robotics backend ran on an ASUS ROG Zephyrus Duo workstation with an AMD Ryzen~9 CPU and an NVIDIA RTX~4090 GPU, under Ubuntu~24 with ROS~2 Jazzy. Communication between headsets and backend was provided through a local Wi-Fi~7 network.

The robot team consisted of three fully simulated Nova Carter mobile robots in NVIDIA Isaac Sim. Robot state, navigation, and sensor streams were exposed to HORUS through the same ROS~2-based bridge used by the runtime, including navigation commands, pose updates, and live camera feeds. This preserved the same supervisory-control and teleoperation workflow used by the application in deployment while shifting the experiment to a photorealistic and repeatable simulated setting.

Two indoor simulation environments were used: an \emph{office} and a \emph{hospital}. Both required team-level exploration and contained doors, room transitions, and partially explored regions in the occupancy map. The environments differed in search structure: in the hospital, points of interest were located inside rooms rather than in corridors, whereas in the office they could appear in both corridors and offices.

\begin{figure}[t]
    \centering
    \includegraphics[width=\columnwidth]{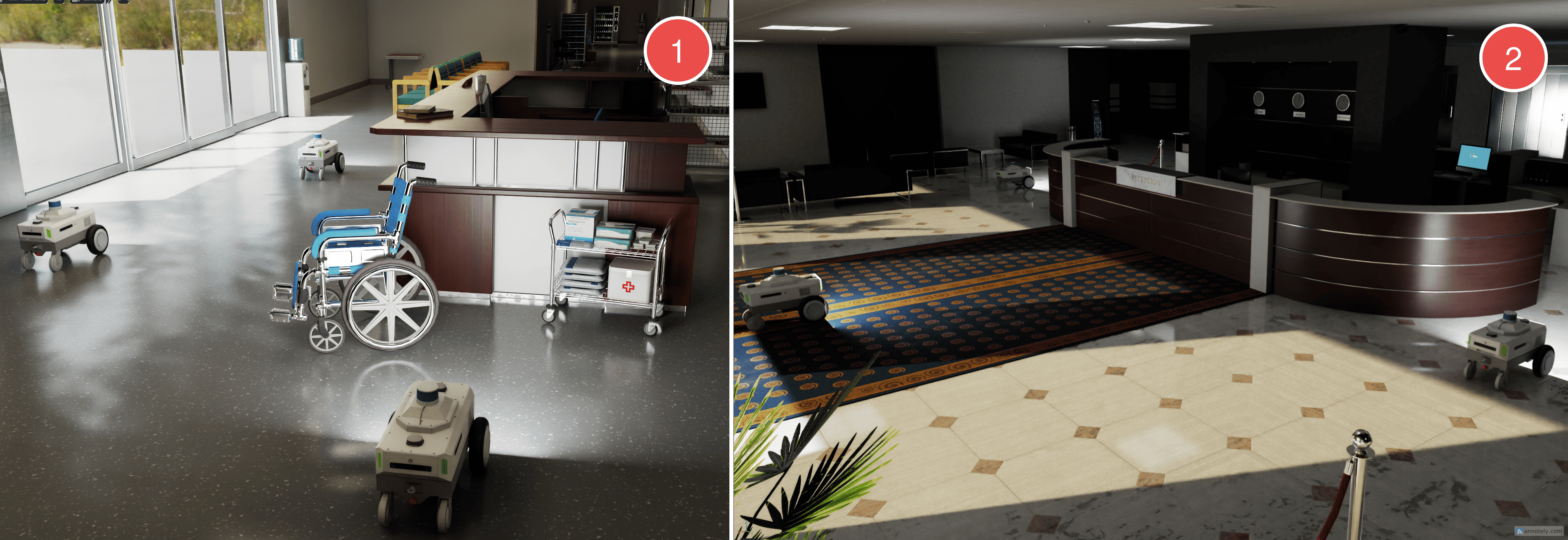}
    \caption{Experiment environment. where 1 is the hospital environment and 2 is the office eenvironment.}
    \label{fig:experimental_design}
\end{figure}

\subsection{Participants and Study Design}

The study involved 36 participants organized into 18 pairs. Each pair experienced both collaborative conditions:
\begin{itemize}
    \item \emph{Co-located shared-workspace}: both participants worked in the same physical room while seeing the same MR workspace aligned in the same physical location;
    \item \emph{Private-workspace}: both participants worked on the same mission, but each used an independently placed local workspace and the pair members were separated into different rooms.
\end{itemize}

The experiment used a counterbalanced crossover design. Each pair performed one trial in each condition, with one trial in the office and the other in the hospital. Condition order and starting environment were counterbalanced across pairs. This kept the session length manageable while still allowing every pair to experience both collaborative modes.

The comparison was intended to isolate the role of shared physical co-location in a realistic deployment of the two modes. Accordingly, co-located trials allowed natural face-to-face verbal coordination, whereas private-workspace trials separated the two participants and did not allow direct verbal exchange.

\subsection{Task}

In each trial, participants explored the environment and searched for AprilTags representing points of interest. Each environment contained 10 AprilTags, and each trial lasted 10 minutes. The objective was therefore to find as many tags as possible within the allotted time.

The pair controlled the same team of three robots in both conditions. No fixed role assignment was imposed: participants were free to self-organize how they used the robots, divide responsibilities, or hand off control dynamically.

The task required participants to combine supervisory planning with direct intervention. Go-to-point commands could be used to cover accessible open areas, while room interiors and tight transitions often required minimap teleoperation or immersive teleoperation. Participants were instructed to attend to the occupancy grid map, the robot camera feeds, and the physical constraints of the robots. In particular, robots could not traverse stairs or pass through doors that were not sufficiently open, and unknown regions were only worth exploring when a reachable path was plausible.

\subsection{Procedure}

Each session began with a guided tutorial and familiarization phase. Participants were introduced to the headset and controllers, the miniature workspace, the Robot Manager, goal assignment, minimap teleoperation, immersive teleoperation, and interpretation of the occupancy grid map. Training duration was not fixed in advance; instead, it continued until both participants indicated that they understood the main interactions and could use the system with confidence. In practice, the tutorial typically lasted approximately 10--15 minutes, and the actual training duration was recorded for later analysis.

After training, each pair completed two timed trials. In the first trial, they were assigned one environment and one collaborative condition according to the counterbalancing schedule, then completed the post-condition questionnaires, specifically the System Usability Scale (SUS) and the NASA Task Load Index (NASA-TLX). To assess collaborative aspects specific to co-located multi-operator mixed reality, we also administered an 8-item custom Collaboration and Shared Workspace questionnaire. These Likert items were adapted from validated constructs of social/co-presence \cite{harms2004networked} and shared mental models \cite{vanrensburg2022psmms}, supplemented with task-specific items on control clarity and handoff. They then performed the second trial in the remaining environment under the other collaborative condition, again followed by the corresponding questionnaires. At the end of the session, participants completed a final preference comparison and a post-experiment simulator sickness questionnaire.

During the study, we recorded objective task outcomes at both the pair and individual levels, together with tutorial duration and post-condition subjective responses on usability, workload, collaboration, condition preference, and simulator sickness. These measures are analyzed in Section~V.

\section{Results}
\label{sec:results}

Statistical analyses were conducted using two-sided Wilcoxon signed-rank tests ($\alpha=0.05$) for within-subject comparisons, unless otherwise specified. Because the experimental task was intrinsically collaborative, objective task performance was analyzed at the pair level ($n=18$). Questionnaire responses were analyzed at the participant level ($n=36$), with the exception of the SUS ($n=35$) due to one incomplete response. Final preference questions were evaluated using exact binomial tests after excluding \emph{No preference} or \emph{No difference} responses. Exploratory correlations relating training time or task contribution to participant backgrounds were assessed using Spearman's $\rho$, and exploratory between-group background comparisons were evaluated via Mann--Whitney tests.

\subsection{System-Level Validation}

All 18 pairs successfully completed both experimental conditions and produced usable data. During the study, the co-located shared-workspace reliably supported the creation, joining, and management of shared sessions. Conversely, the private-workspace mode successfully facilitated the same mission workflow without relying on a common physical anchor. Crucially, the lease arbitration mechanism remained robust in both modes, effectively preventing simultaneous, conflicting interactions with the same robot.

The main practical issue to note concerns headset floor-height calibration. If the Meta headset's room-scale floor estimate is incorrect, the same offset propagates to the vertical placement of the virtual workspace. We observed this twice during the study. In practice, recalibrating the headset boundary before the trial restored the expected workspace height.

\subsection{Human-Subject Study}

\begin{table}[t]
    \centering
    \caption{Main outcomes of the human-subject study. Task performance is analyzed at the pair level ($n=18$); questionnaire measures are analyzed at the participant level ($n=36$), except SUS ($n=35$). Values are mean $\pm$ SD.}
    \label{tab:main_results}
    \begin{tabular}{lccc}
        \hline
        Metric & Co-located & Private & $p$ \\
        \hline
        Tags found (pair) & $3.94 \pm 2.71$ & $4.22 \pm 1.93$ & .633 \\
        SUS & $67.2 \pm 16.3$ & $68.6 \pm 17.9$ & .603 \\
        NASA-TLX (overall) & $3.38 \pm 1.04$ & $3.34 \pm 0.79$ & .935 \\
        Collaboration mean & $5.29 \pm 1.44$ & $4.09 \pm 1.53$ & $<.001$ \\
        \hline
    \end{tabular}
\end{table}

\begin{figure}[t]
    \centering
    \includegraphics[width=\columnwidth]{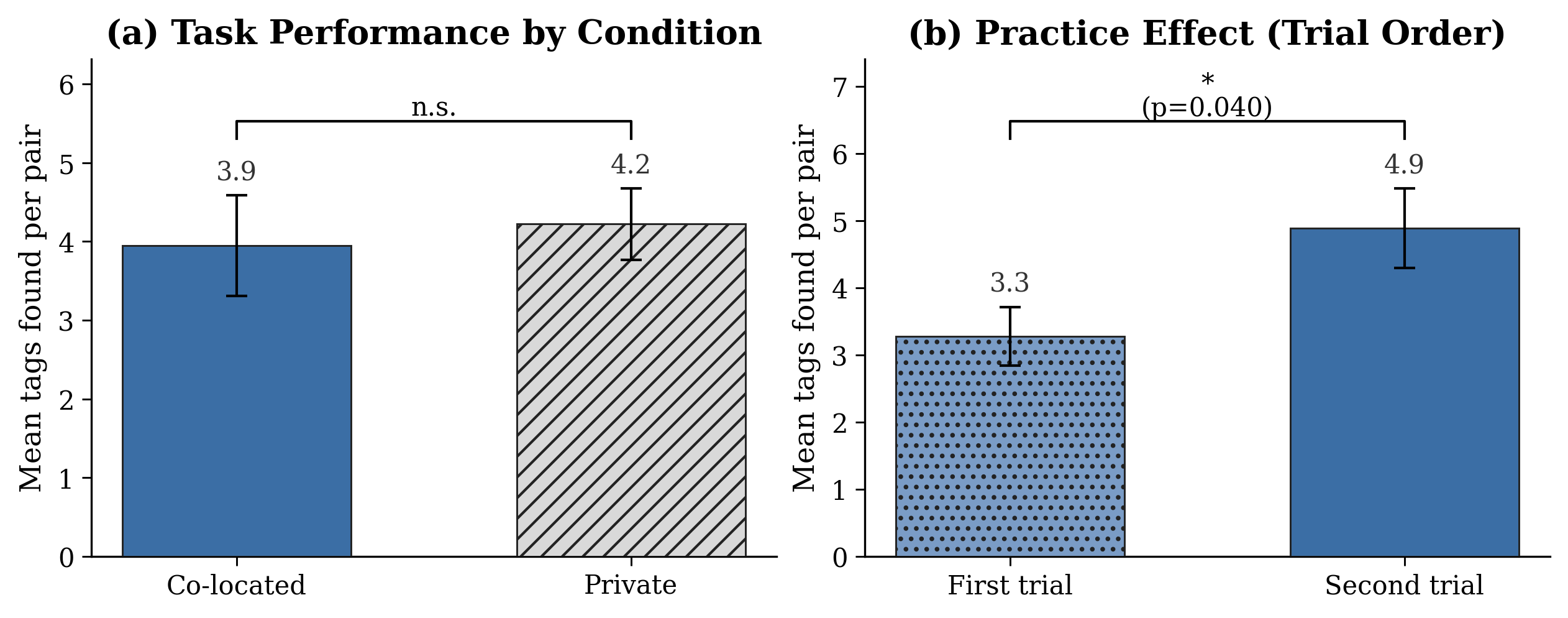}
    \caption{Objective task performance. Left: pair-level AprilTag detections in co-located and private-workspace modes. Right: pair-level detections in the first and second trial, showing the practice effect across the session.}
    \label{fig:performance_results}
\end{figure}

\subsubsection{Task performance}
The primary behavioral metric was the number of AprilTags located per 10-minute trial. At the pair level, the co-located shared-workspace did not significantly outperform the private-workspace condition. Pairs found an average of $3.94 \pm 2.71$ tags in the co-located mode and $4.22 \pm 1.93$ in the private mode, yielding no significant difference ($W=66.5$, $p=.633$). A secondary analysis of individual participant contributions reflected this same parity ($1.97 \pm 1.75$ vs.\ $2.11 \pm 1.41$ tags, $W=193.0$, $p=.592$).

This confirms the hypothesis that the private-workspace mode enables different users to collaborate in the environment even without direct communication. Both modes fully supported multi-robot task execution. Consequently, differences observed in subsequent collaboration metrics are better attributed to the spatial interaction design than to disparities in basic system capability.

Across trials, however, a significant practice effect emerged. When comparing a pair's first and second trials regardless of the assigned condition, tag detection increased significantly in the second trial ($4.89 \pm 2.52$ vs.\ $3.28 \pm 1.84$; $W=33.5$, $p=.040$). This indicates that operators became more proficient at the search task as the session progressed, though this learning effect did not translate into a measurable performance advantage for either specific collaborative mode.

\begin{figure}[t]
    \centering
    \includegraphics[width=\columnwidth]{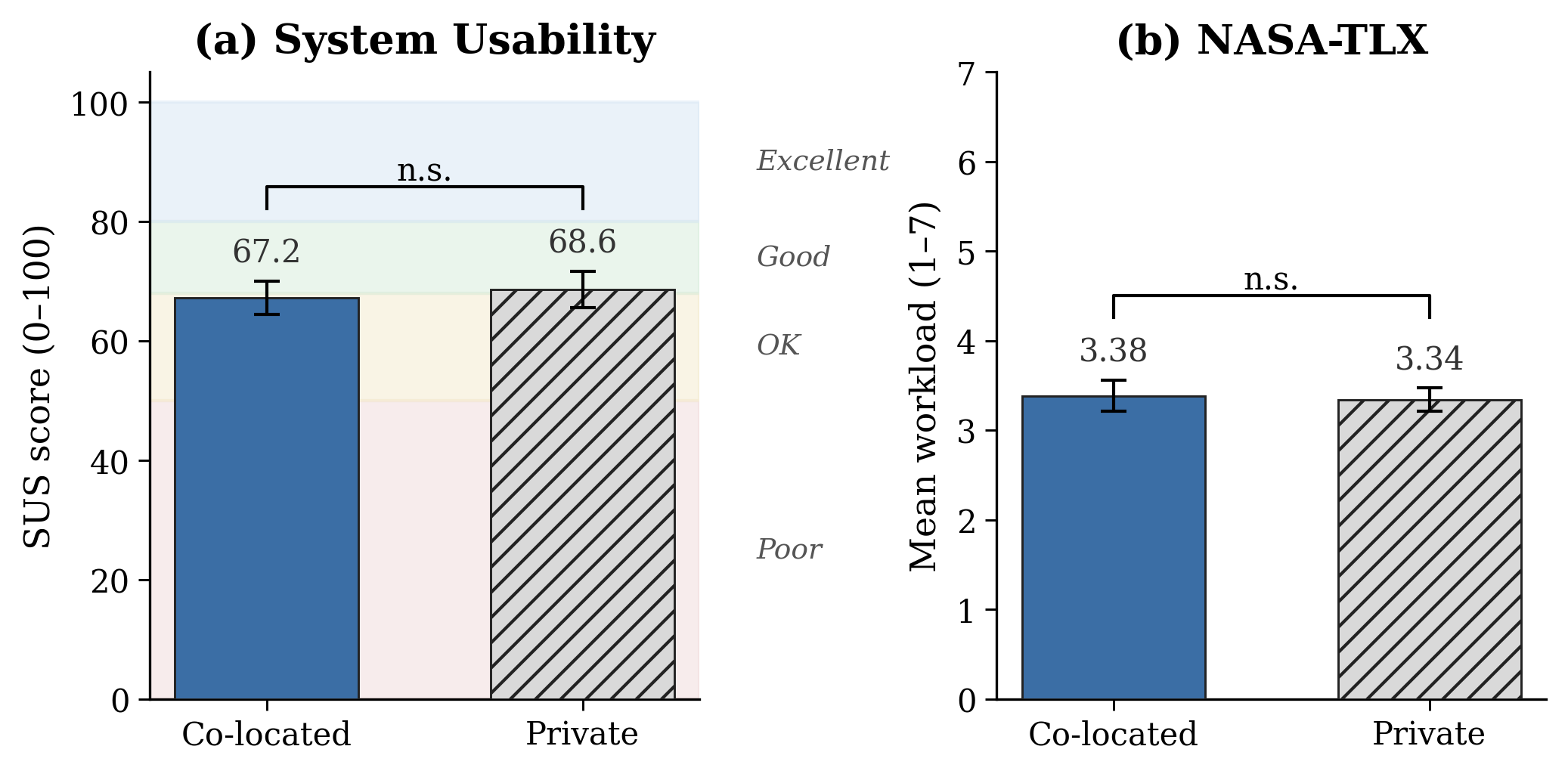}
    \caption{Composite questionnaire outcomes by condition. The co-located and private-workspace modes produced similar usability and workload scores, but the co-located mode yielded markedly higher collaboration ratings.}
    \label{fig:composite_results}
\end{figure}

\subsubsection{Usability, workload, and comfort}
Perceived usability was highly comparable across both collaborative architectures. SUS scores averaged $67.2 \pm 16.3$ in the co-located condition and $68.6 \pm 17.9$ in the private-workspace condition, with no statistical difference ($W=221.5$, $p=.603$).

Overall workload, assessed via the NASA-TLX, was similarly balanced between conditions ($3.38 \pm 1.04$ vs.\ $3.34 \pm 0.79$, $W=310.0$, $p=.935$). At the subscale level, no single workload dimension reached statistical significance. The most notable trend occurred in Mental Demand, which was lower in the co-located mode ($3.42 \pm 1.59$) than in the private mode ($3.89 \pm 1.49$), approaching significance ($W=215.5$, $p=.074$). This trend aligns with the hypothesis that a shared physical workspace may reduce some of the mental effort required to internally model a teammate's ongoing activities. Simulator sickness remained negligible across the experiment ($p=.656$), confirming that the full session did not produce measurable discomfort.

\begin{figure}[t]
    \centering
    \includegraphics[width=\columnwidth]{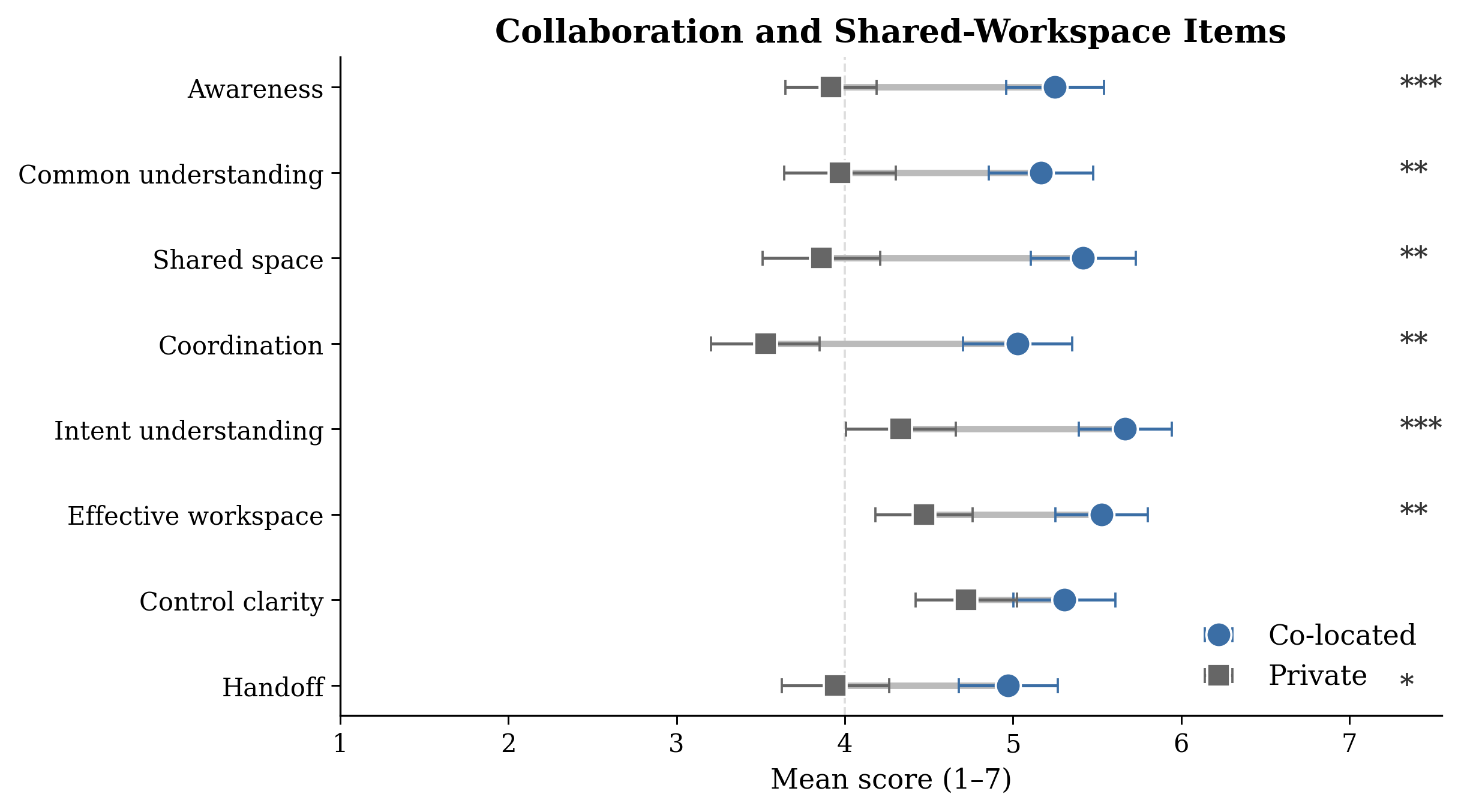}
    \caption{Item-level collaboration ratings. Co-located shared-workspace operation improved awareness of teammate actions, common understanding, perceived shared space, coordination, intent understanding, perceived collaborative effectiveness, and handoff ease.}
    \label{fig:collaboration_items}
\end{figure}

\subsubsection{Collaboration and operator preference}
The most pronounced condition effects emerged in the subjective collaboration assessments. Averaged across all eight items of the custom Collaboration and Shared Workspace questionnaire, the co-located shared-workspace scored significantly higher than the private-workspace ($5.29 \pm 1.44$ vs.\ $4.09 \pm 1.53$, $W=90.5$, $p<.001$). Therefore, despite equal raw task yields, participants found the co-located mode substantially more conducive to effective teamwork.

This advantage was consistent across nearly all individual collaboration items. Operating in the co-located mode significantly enhanced participants' awareness of their teammate's actions ($5.25$ vs.\ $3.92$, $p<.001$), their common understanding of the task state ($5.17$ vs.\ $3.97$, $p=.006$), and their feeling of sharing a physical space ($5.42$ vs.\ $3.86$, $p=.001$). The shared physical anchor also directly facilitated intervention mechanics: participants reported greater ease in coordinating actions ($5.03$ vs.\ $3.53$, $p=.008$), understanding teammate intentions ($5.67$ vs.\ $4.33$, $p<.001$), executing control handoffs ($4.97$ vs.\ $3.94$, $p=.021$), and rated the workspace as generally more effective for collaboration ($5.53$ vs.\ $4.47$, $p=.002$). The only item lacking a significant difference concerned the clarity of which operator currently controlled a robot ($5.31$ vs.\ $4.72$, $p=.115$). This implies that the UI's explicit lease state was legible in both modes, even while the broader collaborative experience strongly favored co-location.

Post-experiment preference queries converged with these ratings. Among those expressing an overall preference, 21 out of 29 participants favored the co-located condition over the private-workspace condition ($p=.024$, exact binomial test). When asked specifically which condition made collaboration easier, the preference was even more decisive, with 27 of 33 favoring co-location ($p<.001$). Furthermore, 21 of 29 reported that control handoffs were clearer ($p=.024$), and 21 of 29 found it easier to maintain awareness of the mission and robot team ($p=.024$) in the co-located mode.

\subsubsection{Learnability and training time}
Participants completed the guided tutorial in an average of $15.44 \pm 5.35$ minutes. To evaluate system learnability, we conducted an exploratory analysis correlating tutorial duration with participants' pre-study backgrounds.

Training time showed no significant association with age ($\rho=-0.12,\ p=.489$) or lifetime MR/VR headset exposure ($\rho=0.01,\ p=.975$). Instead, the strongest independent trends pointing toward faster familiarization were linked to general control proficiency: controller familiarity ($\rho=-0.31,\ p=.065$), prior teleoperation experience ($\rho=-0.30,\ p=.076$), and prior multi-robot supervision ($\rho=-0.30,\ p=.072$).

A composite metric combining these interactive-control variables revealed a significant negative correlation with tutorial duration ($\rho=-0.48,\ p=.003$), indicating that users with a stronger foundation in generalized remote control mastered the interface faster than those relying on prior VR exposure alone. Importantly, tutorial duration did not significantly correlate with first-condition individual tag yield ($\rho=-0.23,\ p=.169$). This suggests that regardless of the time required to learn the interface, participants operated the system with comparable efficacy once training concluded. In any case, the private-workspace condition allowed users to achieve similar task-success results, demonstrating that the implemented strategy for managing multiple users is effective.

\subsubsection{Effect of prior experience on task contribution}
While general control skills accelerated initial learning, domain-specific prior robot-control experience was a stronger predictor of mission success. Participants possessing any prior teleoperation experience located significantly more tags overall compared to novices ($4.87$ vs.\ $2.69$ tags; Mann--Whitney $p=.006$). This performance gap was particularly pronounced in the co-located shared-workspace condition ($2.57$ vs.\ $0.92$, $p=.001$), whereas the difference in the private-workspace mode was narrower and not statistically significant ($2.30$ vs.\ $1.77$, $p=.337$).

This pattern held at the pair level. Teams comprising at least one teleoperation-experienced member significantly outperformed entirely novice teams ($8.87$ vs.\ $4.67$ total tags, $p=.049$), again driven predominantly by performance in the co-located mode ($4.60$ vs.\ $0.67$, $p=.014$). Additionally, a pair's average prior multi-robot supervision experience correlated positively with total objective performance (Spearman $\rho=0.48,\ p=.045$).

\section{Discussion}

The main finding of this study is that co-locating the MR workspace changed \emph{how} participants collaborated more than \emph{how many} targets they found. Both collaborative modes supported comparable task execution, showing that HORUS remained operationally effective whether operators worked over a shared physical workspace or through separate private workspaces. Against this stable baseline, the co-located mode produced clear gains in perceived teammate awareness, common understanding, coordination, and handoff clarity. The main value of co-location therefore appears to be improved collaborative fluency rather than increased short-duration task throughput.

This pattern is consistent with the intended role of the shared workspace. In the co-located mode, operators coordinated around the same physicalized miniature environment, while robot availability and control ownership remained explicitly visible through the lease-state cues in the shared workspace. In the private-workspace condition, the same tools were available, but coordination required more deliberate inference of teammate activity. The near-significant reduction in mental demand in the co-located condition is consistent with this interpretation.

The learnability analysis complements this result. Participants generally became operational after a short tutorial, and training time was more closely related to control-oriented experience than to prior MR/VR exposure. This suggests that the interface can be adopted without substantial prior headset familiarity, while still benefiting from prior experience with teleoperation and related control tasks. The additional finding that prior teleoperation experience was associated with stronger task contribution, especially in the co-located mode, points in the same direction: the proposed interface benefits from general remote-control competence, but it does not depend on previous MR expertise.

Taken together, the results support co-located shared workspaces as a strong design choice for collaborative MR robot supervision. At the same time, the comparable task performance observed in the private-workspace condition suggests that the two modes are complementary rather than mutually exclusive. A practical implication is that future multi-operator MR systems may benefit from supporting both: co-located shared workspaces for tightly coordinated local teams, and private workspaces for collaborators who need to participate without sharing the same physical anchor.

\section{Conclusion}

This paper presented a multi-operator mixed-reality interface for collaborative multi-robot control and coordination. Extending HORUS with collaborative session support and per-robot control leases enabled multiple operators to supervise, task, and teleoperate the same robot team while preventing conflicting interventions. The study compared two collaborative modes: a co-located shared workspace and a private-workspace alternative. Both supported effective task execution, but the co-located shared workspace significantly improved perceived collaboration, shared understanding, and handoff clarity, and was the preferred mode for joint operation. These results indicate that physically co-locating the MR workspace improves how operators coordinate, even when the underlying control tools remain unchanged. More broadly, they suggest that shared spatial references are a useful design principle for collaborative MR command interfaces for robot teams.

\begingroup
\sloppy
\hbadness=10000
\bibliographystyle{IEEEtran}
\bibliography{references}
\endgroup

\end{document}